%% file: main.tex
\documentclass[sigconf]{acmart}

\usepackage{amsmath}
\usepackage{multirow}
\usepackage{algorithm}
\usepackage{url}
\usepackage{algorithmic}

\AtBeginDocument{%
	\providecommand\BibTeX{{%
			\normalfont B\kern-0.5em{\scshape i\kern-0.25em b}\kern-0.8em\TeX}}}

\copyrightyear{2021}
\acmYear{2021}
\setcopyright{acmcopyright}
\acmConference[CIKM '21] {Proceedings of the 30th ACM International Conference on Information and Knowledge Management}{November 1--5, 2021}{Virtual Event, Australia.}
\acmBooktitle{Proceedings of the 30th ACM Int'l Conf. on Information and Knowledge Management (CIKM '21), November 1--5, 2021, Virtual Event, Australia}
\acmPrice{15.00}
\acmISBN{978-1-4503-8446-9/21/11}
\acmDOI{10.1145/3459637.3482331}

\begin{document}
\fancyhead{}

\title{Conditional Graph Attention Networks for Distilling \\ and Refining Knowledge Graphs in Recommendation}

\author{Ke Tu}
\affiliation{%
  \institution{Ant Group}}
\email{tuke.tk@antgroup.com}

\author{Peng Cui}
\affiliation{%
  \institution{Tsinghua University}}
\email{cuip@mail.tsinghua.edu.cn}

\author{Daixin Wang}
\affiliation{%
  \institution{Ant Group}}
\email{daixin.wdx@antgroup.com}

\author{Zhiqiang Zhang}
\affiliation{%
  \institution{Ant Group}}
\email{lingyao.zzq@antgroup.com}

\author{Jun Zhou}
\authornote{Corresponding author}
\affiliation{%
  \institution{Ant Group}}
\email{jun.zhoujun@antgroup.com}

\author{Yuan Qi}
\affiliation{%
  \institution{Ant Group}}
\email{yuan.qi@antgroup.com}

\author{Wenwu Zhu}
\affiliation{%
  \institution{Tsinghua University}}
\email{wwzhu@tsinghua.edu.cn}


\begin{abstract}
	Knowledge graph is generally incorporated into recommender systems to improve overall performance. Due to the generalization and scale of the knowledge graph, most knowledge relationships are not helpful for a target user-item prediction. To exploit the knowledge graph to capture target-specific knowledge relationships in recommender systems, we need to distill the knowledge graph to reserve the useful information and refine the knowledge to capture the users' preferences. To address the issues, we propose \emph{Knowledge-aware Conditional Attention Networks} (KCAN), which is an end-to-end model to incorporate knowledge graph into a recommender system. Specifically, we use a knowledge-aware attention propagation manner to obtain the node representation first, which captures the global semantic similarity on the user-item network and the knowledge graph. Then given a target, i.e., a user-item pair, we automatically distill the knowledge graph into the target-specific subgraph based on the knowledge-aware attention. Afterward, by applying a conditional attention aggregation on the subgraph, we refine the knowledge graph to obtain target-specific node representations. Therefore, we can gain both representability and personalization to achieve overall performance. Experimental results on real-world datasets demonstrate the effectiveness of our framework over the state-of-the-art algorithms.
\end{abstract}


\maketitle

\input{Introduction}
\input{Related_Work}
\input{Framework}
\input{Experiments}
\input{Conclusion}

\section{Acknowledgments}

This work was supported in part by CCF-Ant Group Research Fund.

\balance
\bibliographystyle{ACM-Reference-Format}
\bibliography{ref}

\end{document}

%% file: Introduction.tex
\section{Introduction}
Nowadays, recommender systems~\cite{wang2019explainable,he2017neural,gao2020deep} are widely used in various Internet applications, for example, search engines~\cite{baeza2004query}, video websites~\cite{davidson2010youtube}, and E-commerce~\cite{schafer2001commerce}.
The recommender system aims to find proper items for target users to meet their personalized interests based on the user-item historical network. However, the common critical problem of recommender systems is data sparsity, i.e., the user behaviors or user-item interactions are very limited comparing with the volume of items. Besides, it is hard to recommend items for a new arrival user~\cite{schein2002methods}. To address the limitations, the researchers have proposed to incorporate side information into the recommender systems, such as attributes~\cite{huang2017label}, contexts~\cite{kim2016convolutional}, images~\cite{niu2018neural}.

\begin{figure}
	\centering
	\includegraphics[scale=0.36]{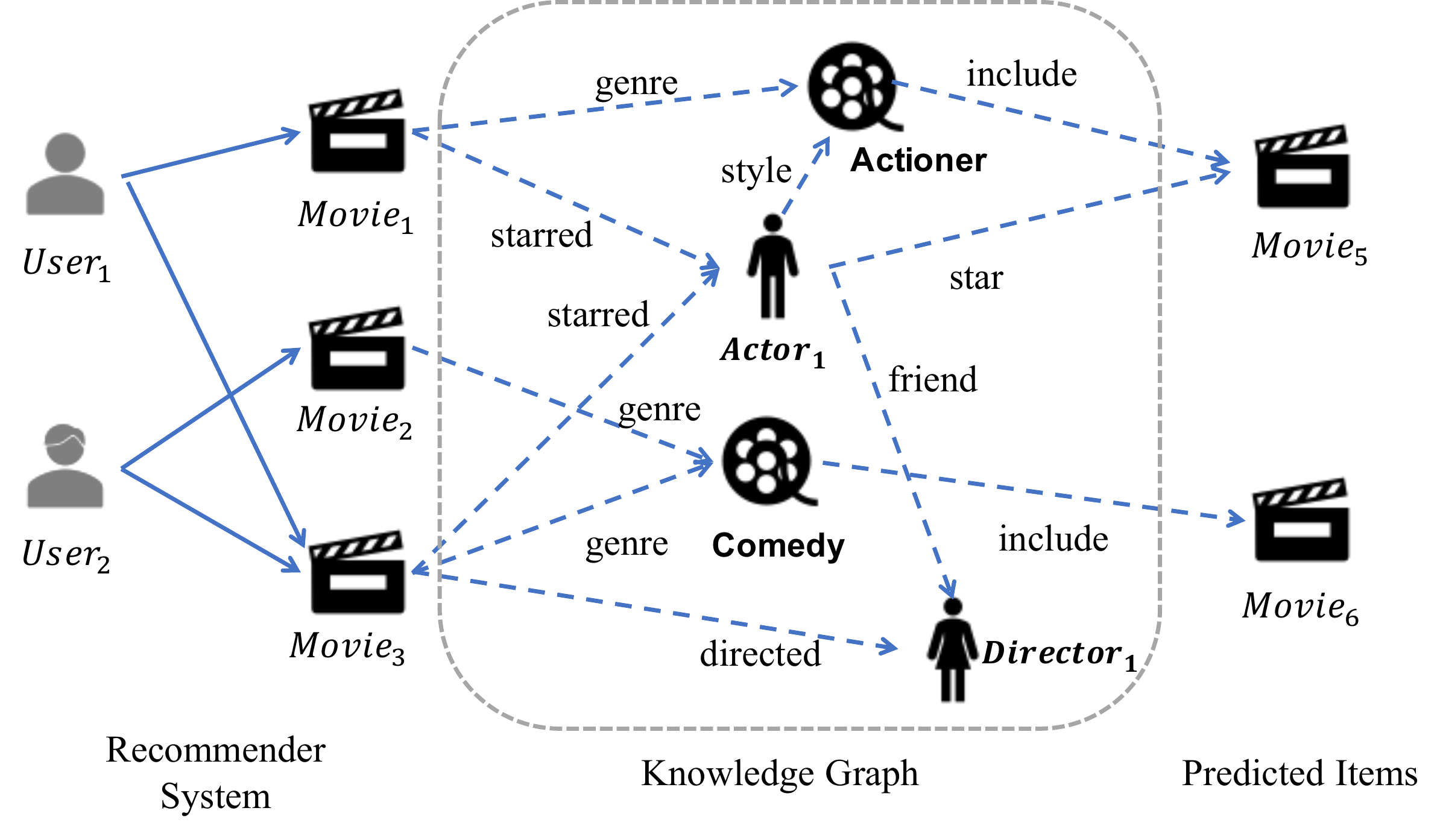}
	\caption{Illustration of Knowledge-aware recommender system.}
	\label{fig:kg}
\end{figure}

Among the various types of side information, knowledge graphs~\cite{wang2017knowledge,dettmers2018convolutional} usually contain more abundant information about the characteristics and connections of nodes. Different from the normal network, the knowledge graph is composed of a set of triplets, i.e., $\textless$head entity, relations, tail entity$\textgreater$. It can describe not only node attributes, such as  $\textless \rm{Movie_1}$, $\rm{Genre}$, $\rm{Actioner}$ $\textgreater$ in Figure~\ref{fig:kg}, but also node relationships, such as $\textless \rm{Actor_1}$, $\rm{Friend}$, $\rm{Director_1} \textgreater$ in Figure~\ref{fig:kg}. Recently, numerous massive knowledge graphs, such as Wordnet~\cite{miller1995wordnet}, Freebase~\cite{bollacker2008freebase} and DBpedia~\cite{auer2007dbpedia}, have been published. These knowledge graphs which describe the facts and common sense can be partly aligned to the nodes of most network applications and be regarded as their side information. They can benefit the recommender systems by introducing relatedness among entities, enriching the entity information, and producing the explainability. However, due to the generalization and scale of the knowledge graph, most knowledge relationships are not helpful for a user-item prediction. To exploit knowledge graph to capture target-specific knowledge relationships in recommender systems~\cite{qin2020survey}, we need to address the following new requirements:
\begin{enumerate}
	\item{\textbf{Knowledge Graph Distillation}:} Knowledge graphs are massive and comprehensive for containing more information. For a specific item recommendation, most knowledge relationships maybe not helpful. Thus learning semantic relationships on the full knowledge graph for a given task is very time-consuming and noisy. Distilling the knowledge graph which gives a small sub-structure related to the target task from the full massive knowledge graph is necessary. It is worth noting that the knowledge graph distillation describes the process of transforming the full knowledge graph into a small concentrated one to capture users' preference accurately, rather than mimicking a pre-trained, larger teacher model by a small student model like knowledge distillation~\cite{hinton2015distilling}.
	\item{\textbf{Knowledge Graph Refinement}:} Personalized recommendation mines user's interests from the past purchasing behaviors. In personalized recommendation with knowledge graph, attention mechanisms are usually used to measure user's preference~\cite{wang2019knowledge,wang2019kgat}. However, they give the same weights of knowledge edges for different target users by edge attentions. For example, in Figure~\ref{fig:kg}, we aim to recommend for targets users $\rm{User_1}$ and $\rm{User_2}$. The  weights of edge $\textless  \rm{Movie_2}, \rm{genre}, \rm{Comedy} \textgreater$ are only based on the nodes $\rm{Movie_2}$ and $\rm{Comedy}$,  and are independent of $\rm{User_1}$ and $\rm{User_2}$. But $\rm{User_1}$ and $\rm{User_2}$ may have different preferences on the genre of the movie. That is to say, the weights of edge <Movie2,genre,Comedy> may be different for $\rm{User_1}$ and $\rm{User_2}$. Thus, for a given target, we should refine the knowledge graph to give different weights for all the knowledge relationships instead of its neighbors.
\end{enumerate}

Therefore, we believe that a good knowledge-aware network learning method should distill and refine the knowledge graphs. Early knowledge graph-aware algorithms are embedding-based models~\cite{wang2019kgat,cao2019unifying}.  They learn entity and relation representations first by knowledge graph embedding algorithms~\cite{wang2014knowledge,dai2020survey} and then incorporate the latent embeddings into the recommender system. The direct way to exploit the knowledge graph in the embedding space fails to solve the distillation and refinement issues and thus harm the performance. The path-based methods~\cite{hu2018leveraging,wang2019explainable,xian2019reinforcement} explore different meta-paths from knowledge graphs to build relationships of two objects, such as $\rm{User}_{1} \stackrel{watch}{\longrightarrow} \rm{Movie}_{1} \stackrel{genre}{\longrightarrow} Actioner \stackrel{include}{\longrightarrow} \rm{Movie}_5$. They distill the knowledge graphs into multiple meta-paths. However, these methods heavily depend on the hand-crafted design of the meta-path. Besides, the multiple paths can not handle the diversity of the users' preference on the knowledge relationships. Recently, some graph convolutional network-based methods~\cite{wang2019kgat,wang2019knowledge} are proposed to propagate information on knowledge graphs. They usually use an attention mechanism~\cite{vaswani2017attention} to produce weights of different neighbors for knowledge graph refinement. However, the attentions are only based on the two nodes in the same edges and are independent with the target nodes.

To overcome the Knowledge Graph Distillation and the Knowledge Graph Refinement issue, we propose a novel model named \emph{Knowledge-aware Conditional Attention Networks} (KCAN). First, we propose a knowledge-aware graph convolutional network to propagate embedding on the knowledge graph by knowledge-aware attention to capture the global similarity of entities like KGAT~\cite{wang2019kgat}. After that,  a subgraph sampling strategy is designed to sample target-specific subgraphs based on the attention weights to distill the knowledge graph. Also, the proposed Local Conditional Subgraph Attention Network (LCSAN) propagates personalized information on the sampled subgraph based on local conditional attention for refining the knowledge graph. In conclusion, the proposed KCAN can effectively distill and refine the knowledge graph at the same time.

It is worthwhile to highlight the following contributions of this paper:
\begin{itemize}
	\item We highlight the importance of distilling and refining knowledge graphs in recommender systems and propose to use a conditional attention mechanism on target-specific subgraphs to capture user preference.
	\item We propose a novel framework named \emph{Knowledge-aware Conditional Attention Networks} (KCAN) to incorporate the knowledge graph into the recommendation. In particular, the \emph{Knowledge-aware Graph Convolutional Network} (KAGCN) layer and target-specific subgraph sampling are designed with the Knowledge Graph Distillation issue in mind. Moreover, the Knowledge Graph Refinement issue is addressed by the \emph{Local Conditional Subgraph Attention Network} (LCSAN) layer.
	\item Extensive experiments on real-world scenarios are conducted to demonstrate the effectiveness of our framework over several state-of-the-art methods.
\end{itemize}

The rest of the paper is structured as follows. We first give a brief review of the related works in section 2. Then we formally define the solved problems and introduce the details of our proposed model in section 3. In section 4, we report the experimental results. Finally, we give a conclusion in section 5. 

\begin{figure*}
	\centering
	\includegraphics[scale=0.33]{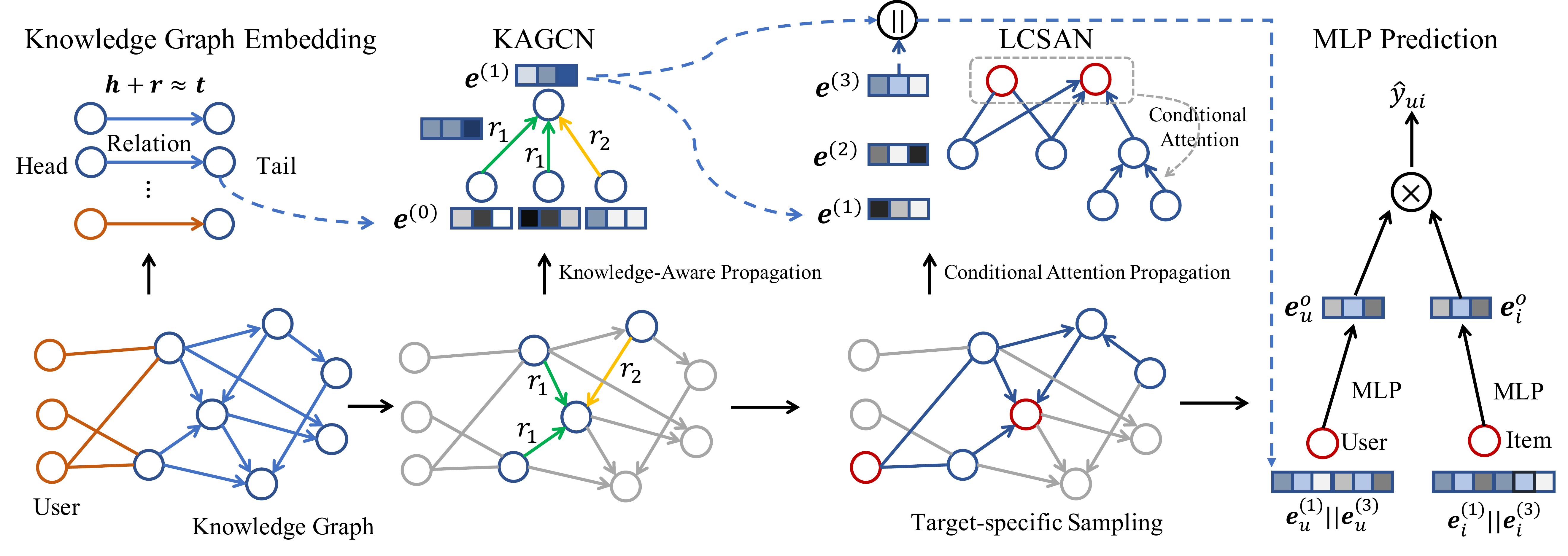}
	\caption{The framework of the proposed KGAN. The framework is composed of four modules: Knowledge Graph Embedding layer, Knowledge Graph Distillation module (KAGCN layer and Target-sepcific Sampling), Knowledge Graph Refinement module (LCSAN layer) and Multi-layer Perceptron (MLP) two-tower prediction layer.}
	\label{fig:framework}
\end{figure*}

%% file: Related_Work.tex
\section{Related Work}
\subsection{Knowledge Graph Embedding}
Knowledge Graph Embedding (KGE)~\cite{wang2017knowledge} aims to learn latent representations for all components of the knowledge graph including entities and relations. Then the low-rank embeddings which preserve the inherent structure and knowledge graph can be used in the downstream tasks such as knowledge graph completion and recommendation. The KGE methods can be roughly divided into two classes: translation distance based models~\cite{bordes2013translating,lin2015learning} and semantic matching based models~\cite{nickel2011three,yang2014embedding,trouillon2016complex}. The translation distance based models exploit distance-based scoring functions. The existing relationships in the knowledge graph will have higher scores. The scoring functions usually measure the distance of two entities under the space of the relation. For example, TransE~\cite{bordes2013translating} assumes $head + relation = tail$ and use $||head+relation-tail||$ as scoring function on the $(head, relation, tail)$ relationships. TransR~\cite{lin2015learning} first projects the entity representations $head$ and $tail$ into the space specific to relation $r$ and measures the distance between entities $head$ and $tail$ under that space. The semantic matching models exploit similarity scoring functions. For example, RESCAL~\cite{nickel2011three} uses a bilinear function to learn the similarity of entities.
For the recommendation task, the main issue is how to introduce the representations of entities and relations into the recommendation system for enriching the item information.

\subsection{Graph Convolutional Network}
The graph convolutional network (GCN)~\cite{kipf2017semi,zhang2020deep} has been proposed to process network data in an end-to-end manner. Earlier works define the graph convolutional operations in the spectral domain. Bruna et al.~\cite{defferrard2016convolutional} are inspired by the graph signal process~\cite{shuman2013emerging} and define the convolution in the Fourier spectral domain.
Kipf et al.~\cite{kipf2017semi} propose to use a first-order approximation to deal with the complexity issue. Recently, lots of non-spectral GCN, which directly define convolution on the graph, have been proposed in the spatial domain. GraphSage~\cite{hamilton2017inductive} samples a fixed-size neighborhood of each node and then aggregates over it. GAT~\cite{velickovic2018graph} introduces a self-attention strategy to specify different weights to different nodes in a neighborhood. However, these works are designed for homogeneous graphs instead of knowledge graphs. Our method is conceptually inspired by GCN. Similarly, Schlichtkrull et al.~\cite{schlichtkrull2018modeling} and Wang et al.~\cite{wang2019knowledge} also propose to apply GCN and GAT into the knowledge graph. They first applies knowledge graph embedding~\cite{wang2017knowledge} methods to obtain representation for entities. Then they propagate the representation over the user-item bipartite graph and knowledge graph to collect the high-order information. Besides, KGCN-LS~\cite{wang2019knowledge2} extends the previous knowledge based GCN method and adds a label smoothness mechanism to propagates the user interaction labels on the knowledge graph. However, the major difference between our work and these methods is that our model consider the knowledge graph distillation and the knowledge graph refinement whiling learning representations.

\subsection{Knowledge-aware Recommendation}
In general, existing knowledge graph-aware algorithms~\cite{qin2020survey,dai2020survey} can be roughly categorized into two types:

(1) Embedding-based methods~\cite{zhang2016collaborative,xin2019relational}. These methods learn entity and relation representations first by knowledge graph embedding algorithms and then incorporate the latent embeddings into the original network. CKE~\cite{zhang2016collaborative} combines collaborative filtering with knowledge graph embedding. Cao et al.~\cite{cao2019unifying} use the knowledge graph to augment the modeling of user-item interaction while completing the missing facts in the knowledge graph based on the enhanced user-item modeling at the same time. Nevertheless, these methods only use knowledge graph embedding as regularization and lose the rich topology structure of the knowledge graph. Additionally, Wang et al.~\cite{wang2019kgat} use graph convolutional network to explicitly model the high-order relations by recursive propagation in the knowledge graph. But all these methods usually have no constraint between the target users and entities and thus hurt the modeling of user preferences. 

(2) Path-based methods. These methods~\cite{wang2019explainable,wang2019heterogeneous,hu2018leveraging} regard the knowledge graph as a heterogeneous information network~\cite{yu2014personalized} and explore different paths from knowledge graphs and measure node relationships by meta-path-based similarity~\cite{sun2011pathsim}. Besides, some methods leverage the meta-path based random walk to learn better entities representation, such as HIN2Vec~\cite{fu2017hin2vec} and metapath2vec~\cite{dong2017metapath2vec}. Defining effective meta-paths usually requires domain knowledge, and it can not generalize to a new dataset.
Recently, some works~\cite{xian2019reinforcement,song2019explainable} propose to apply reinforcement learning to choose meta-paths while training automatically.
However, characterizing the user-item similarity by separate paths may lead to information loss. Our works propose to use target-specific subgraphs to distill the knowledge graph. Xiao et al.~\cite{sha2019attentive} also use subgraphs to distill the knowledge graph. But it pre-computes all user-item subgraphs by the knowledge graph and just learns on these local subgraphs. In such a way, the semantic similarity between entities on the whole knowledge graph is missing. Our model can not only preserve the global semantic similarity but also dynamically generates target-specific subgraphs to distill and refine knowledge graph for inferring local user preference accurately.

%% file: Framework.tex
\section{The Proposed Framework}

In this section, we will introduce the proposed \emph{Knowledge-aware Conditional Attention Networks} (KCAN). The framework is shown in Figure \ref{fig:framework}. The model is composed of four modules: (1) Knowledge Graph Embedding layer. It learns representations for each entity and relation by a traditional knowledge graph embedding method TransH. (2) Knowledge Graph Distillation, which propagates embedding with a knowledge-aware attention mechanism and uses a target-sampling strategy to distill the knowledge graph. (3) Local Conditional Subgraph Attention Network (LCSAN), which propagates personalized information on the subgraph based on local conditional attention to refine the knowledge graph. (4) Multi-layer Perceptron (MLP) two-tower prediction layer, which combines the embeddings from the above two layers and predicts the final results with a non-linear layer.

\subsection{Notations and Definitions}
Let $G = (V, E)$ denotes a network, where $V$ is the set of nodes and $E \subseteq V \times V$ is the set of edges. For a node $v \in V$, $\mathcal{N}(v) = \{u|(v, u) \in E\}$ is the set of its neighbors. In our setting, we have an external knowledge graph $\mathcal{G} = \{(h, r, t) | h \in \mathcal{E}, r \in \mathcal{R}, t \in \mathcal{E}\}$, where $h$, $r$ and $t$ denotes the head entity, relation and tail entity of a knowledge graph triple. $\mathcal{E}$ and $\mathcal{R}$ are the set of entities and relations of knowledge graph $\mathcal{G}$. Actually, our model can be used in any graph tasks with a side knowledge graph. The target sets $\mathcal{T}$ rely on the goal of the graph task. For example, in node classification, the target set is a single node $\mathcal{T}=\{v\}$. For recommendation task, given a bipartite user-item graph, the goal is to predict the existence or similarity $y_{uv}$ of a targeted user-item pair $\mathcal{T}=\{u, v\}$. For drug reactions prediction task, the target set is all the drugs which leads reaction together $\mathcal{T}=\{d_1,.., d_n\}$. In our paper, we only discuss recommendation task with bipartite user-item graph $G$ and an external knowledge graph $\mathcal{G}$. A node in network $G$ may also be an entity in knowledge graph $\mathcal{G}$. In general, we use entity to refer all nodes in both $G$ and $\mathcal{G}$. For an entity $v$, the neighborhood of $v$ in knowledge graph is denoted as $\mathcal{N}(v) = \{(r, v') | (v, r, v') \in \mathcal{G}\}$ and we mark the K-hop neighborhood of entity $v$ as $S_K(v)$. We mark the embedding of entity $h$ in $k$-th layer as $\mathbf{e}_h^{(k)} \in \mathbb{R}^{F_k}$ with  $k=0,1,2,...,K$. For simplify, we mark $\mathbf{e}_h^{(0)}$ as $\mathbf{e}_h$.

\subsection{Knowledge Graph Distillation}
To distill the whole knowledge graph, we must recognize the importance of the knowledge relationships and eliminate the useless relationships. For this purpose, we first vectorize the entities by knowledge graph embedding.

\subsubsection{Knowledge Graph Embedding}
Knowledge graph embedding~\cite{wang2014knowledge,ji2015knowledge,dai2020survey} is an effective way to learn entity and relation representations while preserving the topology structure. Since we regard the click relation in the recommendation as a type of knowledge relationships and a user may click many items, so there are a lot of one-to-many mappings in the knowledge graph.
Here we use TransH~\cite{wang2014knowledge} which can solve the one-to-many and many-to-one issues. For a given triple $(h, r, t)$, its score or distance is defined as follows:
\begin{equation}
f_r(h, t)=\left\|\mathbf{e}_{h\perp}+\mathbf{d}_{r}-\mathbf{e}_{t\perp}\right\|_{1}^{2},
\end{equation}
where $\mathbf{e}_{h\perp}=\mathbf{e}_h-\mathbf{w}_{r}^{\top} \mathbf{e}_h \mathbf{w}_{r}$ and $\mathbf{e}_{t\perp}=\mathbf{e}_t-\mathbf{w}_{r}^{\top} \mathbf{e}_t \mathbf{w}_{r}$ are the projections of entities embedding $\mathbf{e}_h$ and $\mathbf{e}_t$ on the relation-specific hyperplane $\mathbf{w}_r$, and $\mathbf{d}_r$ is the relation-specific translation vector; $\|\cdot\|_1$ is the $L1$-norm.
The lower score of $f_r(h, t)$ indicates the triplet is more likely to be true and vice versa. By projecting to the relation-specific hyperplane, TransH enables different roles of an entity in different triplets.

The loss of knowledge graph embedding is Bayesian personalized ranking loss~\cite{rendle2009bpr}, which aims to maximize the margin between the positive samples and the negative samples:
\begin{equation}
\label{equ:loss_kg}
\mathcal{L}_{kg} = \sum_{(h, r, t) \in \mathcal{E}} \sum_{(h, r, t') \notin \mathcal{E}} {-\ln\ \sigma(f_{r}(h, t')-f_{r}(h, t))},
\end{equation}
where $\sigma$ is the sigmoid function, $t'$ is uniformly sampled by replacing $t$ with another entity randomly. To increase the representation ability, we also train the model on both the knowledge graph $\mathcal{G}$ and the user-item network $G$ with this loss function. For this purpose, we treat all edges in network $G$ with the same type $r_G$. Then the network $G$ can be also seen as a knowledge graph with one and the same relation type. 

\subsubsection{Knowledge-Aware Graph Covolutional Network}
\label{KAGCN}
However, the knowledge graph embedding only focuses on the separate knowledge triples, thus it fails to capture the high-order similarity between entities.
Inspired by KGAT~\cite{wang2019kgat} and KGNN~\cite{lin2020kgnn}, we present to capture the high-order similarity of entities by aggregating the knowledge embedding with graph convolutional network. To distinguish the influence of different types of relations in the knowledge graph, we propose to use a knowledge-aware attention mechanism over the propagation:
\begin{equation}
\mathbf{e}_{\mathcal{N}(v)}=\sum_{(r, t) \in \mathcal{N}(v)} \pi_r(v, t) \mathbf{e}_{t},
\end{equation}
where $\pi_r(v, t)$ is relation-specific attention coefficient. The attention measures the importance of the entity $v$ to $t$ under the relation $r$-specific space. When the entity $t$ is closer to entity $v$ under the relation $r$-specific space, the more information should be propagated. Motivated by this, we define knowledge-aware attention as follows:
\begin{equation}
\label{equ:katt}
\hat{\pi}_r(v, t) =  {\rm softmax}(\cos(\mathbf{e}_{v\perp}+\mathbf{d}_r, \mathbf{e}_{t\perp})),
\end{equation}
where $\cos (\mathbf{x}, \mathbf{y})=\frac{\mathbf{x}^{\mathrm{T}} \mathbf{y}}{\|\mathbf{x}\|_{2} \|\mathbf{y}\|_{2}}$ is the cosine similarity. In this way, it will propagate more information for closer entities.

Finally, we combine the entity representation $\mathbf{e}_v$ and the neighborhood representation $e_{\mathcal{N}(v)}$ and update the entities representation as $\mathbf{e}_v^{(1)}$. For simplicity, we concatenate two representations and add a nonlinear transformation like GraphSAGE aggregator~\cite{hamilton2017inductive}:
\begin{align}
\label{equ:e1}
\mathbf{e}_v^{(1)} &= {\rm AGG}(\mathbf{e}_v, \mathbf{e}_{\mathcal{N}}(v)) \\
&= {\rm LeakyReLU}(\mathbf{W}^{(1)}(\mathbf{e}_v || \mathbf{e}_{\mathcal{N}}(v))+\mathbf{b}^{(1)}),
\end{align}
where $||$ is the concatenation operation, and LeakyReLU is the activation function. $\mathbf{W}^{(1)}, \mathbf{b}^{(1)}$ are the trainable weights. By this knowledge-aware propagation mechanism, we can preserve the global similarity between entities after several iterations.

\subsubsection{Target-specific Sampling}
For a given user-item target $(u, v)$, the attentions of KAGCN rely on the knowledge edges and do not concern with the targets. To capture the local influence of the targets,  we exploit subgraphs instead of the whole enormous knowledge graph. Compared to the meta-path or random-walk based model which uses path to capture the locality, the subgraph can contain diversity. 
A target set is marked as $\mathcal{T} = \{v_1,...,v_k\}$ which $k=2$ and $\mathcal{T}$ is a user-item pair. Due to the locality of the users' preference, we set the receptive field of each node $v$ as their K-hop neighborhood $S_K(v)$. In a real-world knowledge graph, the number of neighbors may vary significantly over all entities, and some entities may have a huge number of neighbors. To distill the knowledge graph and reduce the training time, we sample a fixed-size set of neighbors for each entity instead of the full neighborhood. We set the sampled size as $M$. The traditional way is to sample neighbors uniformly like GraphSAGE~\cite{hamilton2017inductive}. However, the knowledge graph is usually noisy and full of target-independent information. Additionally, considering that we have obtained the global attention $\pi_r(v, t)$, which measures the similarity of entities, we sample neighbors with their attention score $\pi_r(v, t)$ as the sampled probability. After getting the sampled $K$-hop neighborhood as $\hat{S}_K(v)$ for each node $v$, we can obtain the receptive field of the target set by merging them:
\begin{equation}
	\hat{S}_K(\mathcal{T}) = \cup_{v \in \mathcal{T}} \hat{S}_K(v).
\end{equation}

\subsection{Knowledge Graph Refinement}
The knowledge graph distillation module distills the knowledge graph to a target-specific subgraph as the receptive field of the target set. In this part, we re-weight the knowledge to refine the knowledge graph.
For that, we propagate entities' information on the distilled target-specific subgraph. To capture the preference of the target to the knowledge relationships, we propose to use a conditional attention mechanism over propagation to refine the subgraph as follows:

\begin{equation}
	\mathbf{e}^{(1)}_{\mathcal{N}^\mathcal{T}(v)} = \sum_{(r, t) \in \mathcal{N}^\mathcal{T}(v)}\alpha(v, r, t | \mathcal{T}) \mathbf{e}^{(1)}_t,
\end{equation}
where $\alpha(v, r, t | \mathcal{T})$ is the conditional attention which relies on target $\mathcal{T}$, and $\mathcal{N}^\mathcal{T}(v)$ is the neighbors of the entity $v$ in the receptive field of the target set $\hat{S}_k(\mathcal{T})$.

To better refine the subgraph based on the target, the conditional attention should contain two aspects: (1) the importance $\alpha_1(v, r, t)$ of the knowledge relationship $(v, r, t)$. This part is independent of the target, and it measures the importance of knowledge relationship itself upon the task. If the knowledge relationship is a noise edge in the task, we should reduce its influence. In the KAGCN part of the section \ref{KAGCN}, we had measured it by the knowledge-aware attention $\hat{\pi}_r(v, t)$. So we can just set $\alpha_1=\hat{\pi}_r(v, t)$ for simplicity. (2) the importance $\alpha_2(t | \mathcal{T})$ of the entity to the target set. This term measures the local preference of the target to the tail entity. The entities which are more similar to the target should be more important for the target. To make the target set and the entity comparable, we calculate the representation of the target set as the concatenation of all contained entities, i.e., $\mathbf{e}_{\mathcal{T}} = \|_{v \in \mathcal{T}} \mathbf{e}^{(1)}_v$. Subsequently, we can measure the importance of the entity to the target set by their representations:
\begin{equation}
	\alpha_2(t | \mathcal{T}) = \mathbf{a}^\mathrm{T}[\mathbf{W}^{(1)}_t \mathbf{e}_{\mathcal{T}} \| \mathbf{W}^{(1)}_e \mathbf{e}^{(1)}_t],
\end{equation}
where $\mathbf{a}$ is a weight vector. $\mathbf{W}^{(1)}_t$ and $\mathbf{W}^{(1)}_e$ are the linear transformation matrices of targets and entities respectively.

Combining the two importance scores, we get the conditional attention:
\begin{equation}
	\label{equ:conatt}
	\alpha(v, r, t | \mathcal{T}) = {\rm softmax}({\rm LeakyReLU}(\alpha_1(v,r,t)*\alpha_2(t|\mathcal{T}))).
\end{equation}
The larger attention indicates the more important of the knowledge related to the target. By the conditional attention, the local knowledge graph is refined into a weighted graph varying with the target. 

Finally, similar to KAGCN, the target-specific entities representations can be obtained as follows:
\begin{align}
	\label{equ:agg2}
	\mathbf{e}_{v | \mathcal{T}}^{(2)} &= {\rm AGG}(\mathbf{e}_v^{(1)}, \mathbf{e}^{(1)}_{\mathcal{N}^\mathcal{T}(v)}) \\
				&={\rm LeakyReLU}(\mathbf{W}^{(2)}(\mathbf{e}_v^{(1)} \| \mathbf{e}^{(1)}_{\mathcal{N}^\mathcal{T}(v)})+\mathbf{b}^{(2)}).
\end{align}

Furthermore, in order to capture high-order preference, we further stack more LCSAN layers. Especially, since the subgraphs are composed of $K$-hop composed neighborhoods, we stack the LCSAN with $K$ times to make sure that the information of each node can be propagated over all the subgraph. In general, the $i$-th LAGCN is as follows:
\begin{equation}
	\label{equ:merge2}
	\mathbf{e}_{v|\mathcal{T}}^{(i+1)} ={ \rm AGG}(\mathbf{e}_{v|\mathcal{T}}^{(i)}, \mathbf{e}^{(i)}_{\mathcal{N}^\mathcal{T}(v)}),
\end{equation}
where $i=2,...K$. We set $K=2$ in all our experiments.

\subsection{Multi-layer Perceptron Prediction}
After conducting the LAGCN, we obtain the target-specific representations $\mathbf{e}_{v|\mathcal{T}}^{(K+1)}$. To increase the representability of the embeddings, we concatenate it with the output representations $\mathbf{e}_v^{(1)}$ of KAGCN as $\mathbf{e}_v^c = [\mathbf{e}_v^{(1)} \| \mathbf{e}_{v|\mathcal{T}}^{(K+1)}]$. In such way, we can preserve both global similarity and local preference effectively. To automatically balance the two parts, we feed the concatenated representations $\mathbf{e}_{v|\mathcal{T}}^o$ into a multi-layer perceptron (MLP) layers to obtain the final output representations:
\begin{equation}
	\mathbf{e}_{v|\mathcal{T}}^o = \mathbf{W}^o([\mathbf{e}_v^{(1)} \| \mathbf{e}_{v|\mathcal{T}}^{(K+1)}])+\mathbf{b}^o,
\end{equation}
where $\mathbf{W}^o$ and $\mathbf{b}^o$ are learnable weights. For recommendation with user-item $(u, i)$ prediction, the predicted score is their inner product of user and item representations with the given target $\mathcal{T}_{ui}=\{u, i\}$:
\begin{equation}
	\label{equ:score}
	p(u, i | \mathcal{T}_{ui}) = \hat{y}_{ui} = \mathbf{e}_{u | \mathcal{T}}^{o\mathrm{T}}\mathbf{e}_{i|\mathcal{T}}^o.
\end{equation}

Since we only observe the positive interactions in the recommender systems, we randomly sample some unobserved relations as the negative samples. To make the learned similarities of the positive interactions are larger than the negative samples, we optimize the model with the Bayesian personalized ranking loss~\cite{rendle2009bpr}:
\begin{equation}
\label{equ:loss_t}
	\mathcal{L}_{T} = \sum_{(u,i) \in E^{+}, (u, j) \notin E^{-}} -\ln\sigma(p(u, j | \mathcal{T}_{uj})-p(u, i | \mathcal{T}_{ui})),
\end{equation}
where $E^{+}$ is the observed edges between user $u$ and item $i$ and $E^{-}$ is the uniformly sampled unobserved edges.

\subsection{Optimization}
\subsubsection*{Objective Function.} 
Combing the loss in Equation \ref{equ:loss_kg} and \ref{equ:loss_t}, we get the total objective function as follows:
\begin{equation}
	\mathcal{L} = \mathcal{L}_{kg}+\mathcal{L}_T+\lambda\|\Theta\|_2^2,
\end{equation}
where $\Theta=\{\ \mathbf{\hat{E}}, \mathbf{W}^{(i)}, \mathbf{b}^{(i)}, \mathbf{W}_t^{(j)}, \mathbf{W}_e^{(j)}, \mathbf{W}^o, \mathbf{b}^o | \forall i \in \{1,2,..,K+1\}, \forall j \in \{1,2,...,K\} \}$ is the set of all parameters, and $\mathbf{\hat{E}}$ is the embeddings of all entities and relations. $\lambda$ is the weight of regularization.
We optimize $\mathcal{L}_{kg}$ and $\mathcal{L}_T$ alternatively and Adam~\cite{kingma2014adam} is used to optimize these parameters.
The detailed training algorithm is shown in Algorithm \ref{alg}.
The learning rate for Adam is initially set to 0.025 at the beginning of the training, and the total epoch number is set as 200.

\begin{algorithm}[t]
	\caption{Training Algorithm for \emph{Knowledge-aware Conditional Attention Networks} (KCAN)}
	\label{alg}
	\begin{algorithmic}[1]
		\REQUIRE User-item network $G$; Knowledge graph $\mathcal{G}$. The configuration $\Theta=\{\ \mathbf{\hat{E}}, \mathbf{W}^{(i)}, \mathbf{b}^{(i)}, \mathbf{W}_t^{(j)}, \mathbf{W}_e^{(j)}, \mathbf{W}^o, \mathbf{b}^o | \forall i \in \{1,2,..,K+1\}, \forall j \in \{1,2,...,K\} \}$.
		\STATE Initialize all configurations
		\FOR{epoch $\gets  0, 1, ..., total\_epoch\_number$}
			\STATE /* Phase I  Knowledge Graph Distillation*/
			\FOR{each triple (h, r, t) in $\mathcal{G}$}
				\STATE Compute the loss of knowledge graph embedding in Equation \ref{equ:loss_kg} and update knowledge graph embedding $e^{(0)} \in \Theta$ by Adam.
				\STATE Compute the knowledge-aware attention $\hat{\pi}_r(v, t)$ in Equation \ref{equ:katt}.
			\ENDFOR
			\STATE Based on Equation \ref{equ:e1}, propagate over the knowledge graph and the network by the knowledge-aware attention $\hat{\pi}_r(v, t)$ to obtain embedding $e^{(1)}$.
			\STATE Based on $\hat{\pi}_r(v, t)$, sample target-specific subgraph.
			\STATE
			\STATE /* Phase II Knowledge Graph Refinement */
			\FOR{each target $\mathcal{T} = \{u, i\}$ in $G$}
				\FOR{$k \gets 1,...,K$}
					\STATE Based on Equation \ref{equ:conatt}, compute the conditional attention $\alpha(v, r, t | \mathcal{T})$.
					\STATE Based on Equation \ref{equ:agg2} and \ref{equ:merge2}, propagate over the target-specific subgraph $\hat{S}_K(\mathcal{T})$ by the conditional attention $\alpha(v, r, t | \mathcal{T})$ to obtain embedding $e^{(k+1)}_{\cdot|\mathcal{T}}$.
				\ENDFOR
				\STATE Based on $e^{(1)}$ and $e^{(K+1)}_{\cdot|\mathcal{T}}$, compute the predicted score $p(u,i|\mathcal{T})$ by Equation \ref{equ:score}.
				\STATE Compute the target loss in Equation \ref{equ:loss_t} and update the configuration $\Theta$ by Adam.
			\ENDFOR
		\ENDFOR
	\end{algorithmic}
\end{algorithm}

\subsubsection*{Training Complexity.}
The time cost of our proposed KCAN mainly comes from two part, KAGCN and LCSAN. The time complexity of KCAN is $O((|E|+|E_{kg}|)*F_1^2)$ where $|E|$ is the number of edges in network $G$, $|E_{kg}|$ is the number of triples in knowledge graph $\mathcal{G}$ and $F_1$ is the dimension of the knowledge graph embedding. For LCSAN, the time complexity is $O(|E_{kg}|*M*\sum_{i=1}^KF_iF_{i+1})$, where $M$ is the fixed-size number of neighbors while sampling subgraphs, and $F_i$ is embedding size of $i$-th layer. We usually set $S$ as 20. So the total complexity of KCAN is $O((|E|+|E_{kg}|)*F_1^2+|E_{kg}|*S*\sum_{i=1}^KF_iF_{i+1})$. Note that the $M$ and $F_i$ are small user-specified constants, the proposed KCAN is linear to the scale of network and knowledge graph. 

%% file: Experiments.tex
\section{Experiments}
In this section, we evaluate our method on three real-world datasets to prove its efficacy. 

We aim to answer the following research questions:
\begin{itemize}
    \item \textbf{RQ1}: What do the knowledge attention and the conditional attention in the proposed model learn? 
	\item \textbf{RQ2}: Does our proposed KCAN outperform the state-of-the-art knowledge-aware methods?
	\item \textbf{RQ3}: How do our proposed KAGCN, LCSAN layers affect the performance of KCAN respectively?
	\item \textbf{RQ4}: How do different choices of hyper-parameters affect the performance of KCAN?
\end{itemize}

\subsection{Datasets and Baselines}
\begin{table}[]
	\caption{Statistics of the datasets.}
	\label{tab:data}
	\centering
	\resizebox{\columnwidth}{!}{%
	\begin{tabular}{cl|rrr}
		\hline
		\multicolumn{1}{l}{} &
		&
		\multicolumn{1}{c}{MovieLens} &
		\multicolumn{1}{c}{Last-FM} &
		\multicolumn{1}{c}{Yelp} \\ \hline
		\multicolumn{1}{c|}{\multirow{3}{*}{Network}} & \#Users        & 6,036     & 23, 566     & 45, 919     \\ \cline{2-2}
		\multicolumn{1}{c|}{}                         & \#Items        & 2,445     & 48, 123     & 45, 538     \\ \cline{2-2}
		\multicolumn{1}{c|}{}                         & \#Interactions & 376,886   & 3, 034, 796 & 1, 185, 068 \\ \hline
		\multicolumn{1}{c|}{\multirow{3}{*}{\begin{tabular}[c]{@{}c@{}}Knowledge\\ Graph\end{tabular}}} &
		\#Entities &
		182,011 &
		58, 266 &
		90, 961 \\ \cline{2-2}
		\multicolumn{1}{c|}{}                         & \#Relations    & 12        & 9           & 42          \\ \cline{2-2}
		\multicolumn{1}{c|}{}                         & \#Triplets     & 1,241,995 & 464, 567    & 1, 853, 704 \\ \hline
	\end{tabular}}
\end{table}

In order to comprehensively evaluate the effectiveness of our proposed method KCAN, we use three public benchmark datasets: MovieLens, Last-FM and Yelp.

\begin{itemize}
	\item \textbf{MovieLens\footnote{\url{https://grouplens.org/datasets/movielens/1m/}}} is a widely used benchmark dataset in movie recommendations. It contains the explicit ratings (ranging from 1-5) on the MovieLens website. We transform the rating into implicit feedback where each entry is marked with 1 indicating that the user has rated the item over a threshold score (4 for this dataset) and otherwise 0. 
	\item \textbf{Last-FM\footnote{\url{https://grouplens.org/datasets/hetrec-2011/}}} is a music listening dataset collected from Last.fm online music systems. The timestamp of the dataset is from Jan, 2015 to June, 2015. To ensure the quality of the dataset, we use the 10-core setting like \cite{wang2019kgat}, i.e., retaining users and items with at least ten interactions.
	\item \textbf{Yelp\footnote{\url{https://www.yelp.com/dataset/challenge}}} is obtained from the 2018 edition of Yelp challenge. The items are local businesses like restaurants and bars. Similarly, we also use 10-core setting on this dataset.
\end{itemize}

Besides, we also need to construct a corresponding knowledge graph from a massive knowledge graph for each dataset. Actually, it is a rule-base rough knowledge graph distill process. Microsoft Satori\footnote{\url{https://searchengineland.com/library/bing/bing-satori}} is used to build a knowledge graph for MovieLens, we first select a subset of triplets whose relation name contains "movie" and the confidence level is greater than 0.9 from the whole knowledge graph. We match the entities with the items by their title names. Similarly, the corresponding knowledge graph is built from Freebase for Last-FM. For Yelp2018, we use the local business information network, for example, category, location, and attribute, as the knowledge graph.

The detailed statistics of the user-item networks and the knowledge graphs are summarized in Table \ref{tab:data}.

We compare our proposed KCAN with four representative types of baselines, including regularization-based (CKE~\cite{zhang2016collaborative}), factorization-based (NMF~\cite{he2017neural}), path-based (RippleNet~\cite{wang2018ripplenet}) and GCN-based (KGAT~\cite{wang2019kgat}).
\begin{itemize}
	\item \textbf{CKE}~\cite{zhang2016collaborative} combines collaborative filtering with the structural knowledge content, the textual content and visual content in an unified framework. In this paper, we implement CKE by combining collaborative filtering and the structural knowledge content. It uses knowledge graph information as regularization to fine tune the collaborative filtering.
	\item \textbf{NMF}~\cite{he2017neural} is a novel factorization machine model for prediction under sparse settings. It deepens factorization machine under the neural network framework for learning higher-order and non-linear feature interactions.
	\item \textbf{RippleNet}~\cite{wang2018ripplenet} combines regularization-based and path-based methods.It stimulates the propagation of user preferences over the set of knowledge entities.
	\item \textbf{KGAT}~\cite{wang2019kgat} introduces graph attention into recommendation and explicitly models the high-order connectivities in knowledge graph in an end-to-end fashion.
\end{itemize}

We uniformly set the embedding size as 16 for all methods. The hidden size of our method and KGAT is set as a tower structure with 16, 8, 8. For our model and RippleNet, the number of hops is set as 2. The dropout rates of NFM, KGAT, and our model are set as $0.1$. The fixed-size number of neighbors while sampling subgraphs is set as 20 for our model. We do grid search from $\{0.01, 0.025, 0.05, 0.1\}$ to tune the learning rate parameter and from $\{10^{-1}, 10^{-2}, 10^{-3}, 10^{-4}, 10^{-5}\}$ for the weights of the $L_2$ normalization. 
All the experiments are conducted on Intel(R) Xeon(R) CPU E5-2699 v4 @ 2.20GHz with GeForce GTX Titan X GPU.

\begin{figure*}
	\centering
	\includegraphics[scale=0.6]{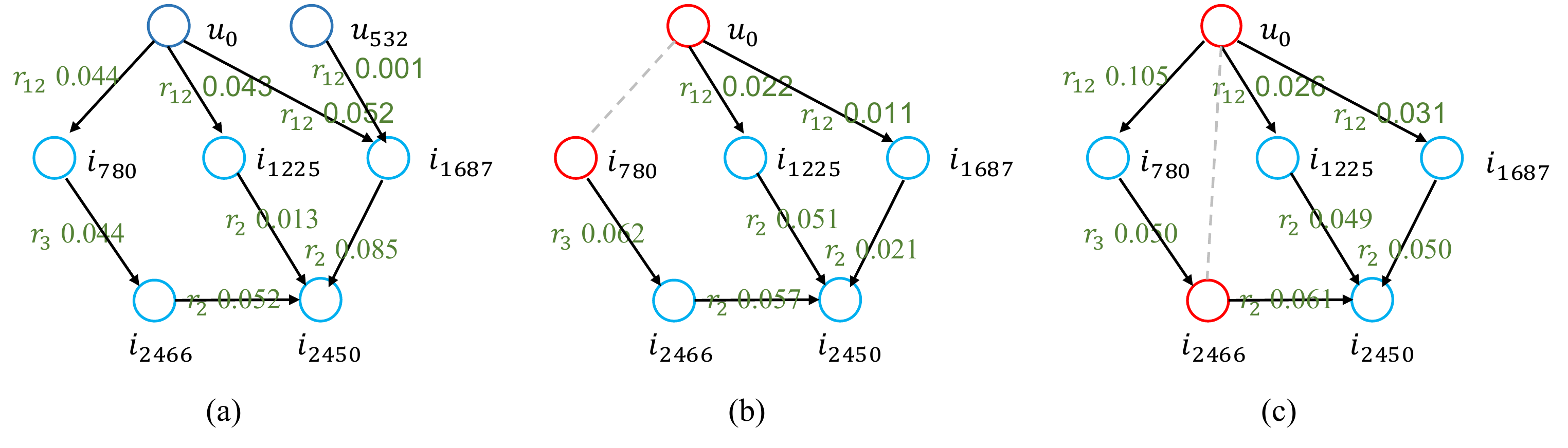}
	\caption{(a) Example from MovieLens dataset.The knowledge attention in the whole knowledge graph and the user-item graph. (b) The conditional attention on the target $(u_0, i_{780})$-specific subgraph. (c) The conditional attention on the target $(u_0, i_{2466})$-specific subgraph. The red nodes are the nodes which needs to be predicted.}
	\label{fig:case_study}
\end{figure*}

\subsection{Case Study (RQ1)}
In this subsection, we show what the knowledge attention and the conditional attention in the proposed model learn on the MovieLens dataset in Figure \ref{fig:case_study}. 

Figure \ref{fig:case_study} (a) shows the global knowledge attention in the Equation \ref{equ:katt}. The attention is only related to knowledge relationships which is the same as the KGAT. From this way, we can infer user preferences. The larger attention means the more important edge. As we can see, the path $u_0 \stackrel{r_{12}}{\longrightarrow} i_{1687} \stackrel{r2}{\longrightarrow} i_{2450}$ has the
highest attention score between nodes $u_0$ and $i_{2450}$. We can explain that the item $i_{2450}$ is recommend to user $u_0$ because the user $u_0$ clicked item $i_{1687}$ which is similar to item $i_{2450}$.
Besides, the edge $u_{532} \stackrel{r_{12}}{\longrightarrow} i_{1687}$ with small weights $0.001$ has less information. We can filter it out to distill the knowledge graph.

The Figure \ref{fig:case_study} (b), (c) show the local attention (Equation \ref{equ:conatt}) with targets $(u_0, i_{780})$, $(u_0, i_{2466})$ respectively. The red nodes are the nodes which needs to be predicted. We refine the target-specific subgraph with different weights by the local attention. We can see that there are different attentions for the same edge $u_{0} \stackrel{r_{12}}{\longrightarrow} i_{1687}$ in the two subgraphs. The edge $u_{0} \stackrel{r_{12}}{\longrightarrow} i_{1687}$ is more important for the prediction of $(u_0, i_{2466})$ than the prediction of $(u_0, i_{780})$. All the baselines can not learn this information which is important for predicting user-item pair.

\subsection{Performance Comparison (RQ2)}
We evaluate our method in two recommendation scenarios. (1) Top-K recommendation. We use the leave-one-out strategy, which has been widely used in previous works~\cite{he2017neural,rendle2009bpr} to evaluate the recommendation performance. For a user, we randomly sample 100 items that are not interacted by the user, ranking the test item among the 100 items. The performance is judged by Hit Ratio@K (Hit@K) and Normalized Discounted Cumulative Gain@K (NDCG@K)~\cite{he2015trirank}. If the true test item ranks in the top $K$ lists, the Hit@K will be one otherwise zero. Compared with Hit@K, NDCG@K pays more attention to the ranking order. The more front position of the true item will have a larger NDCG@K. Then we compute the two metrics for each user and obtain the average score at $K = 10$.  (2) Click-through rate (CTR) prediction. We predict the score of each user-item pair, including positive items and randomly sampled negative items, by the trained model. The evaluation metric in CTR prediction is set as the area under the curve(AUC). The AUC is equivalent to the probability of positive samples are ranked higher than negative samples ~\cite{mason2002areas}.

\begin{table}[]
	\centering
	\caption{Hit@10 and NDCG@10 in top-K recommendation.}
	\label{tab:topk}
	\resizebox{\columnwidth}{!}{%
	\begin{tabular}{c|c|ccccc}
		\hline
		\multicolumn{2}{c|}{Dataset}         & CKE   & NFM   & Ripple & KGAT  & KCAN          \\ \hline
		\multirow{2}{*}{MovieLens} & Hit@10  & 0.293 & 0.384 & 0.607  & 0.467 & \textbf{0.668} \\ \cline{2-2}
		& NDCG@10 & 0.158 & 0.204 & 0.324  & 0.252 & \textbf{0.365} \\ \hline
		\multirow{2}{*}{Last-FM}   & Hit@10  & 0.606 & 0.697 & 0.650  & 0.699 & \textbf{0.771} \\ \cline{2-2}
		& NDCG@10 & 0.383 & 0.421 & 0.367  & 0.437 & \textbf{0.506} \\ \hline
		\multirow{2}{*}{Yelp}      & Hit@10  & 0.741 & 0.775 & 0.721  & 0.799 & \textbf{0.810} \\ \cline{2-2}
		& NDCG@10 & 0.480 & 0.491 & 0.410  & 0.502 & \textbf{0.527} \\ \hline
\end{tabular}}
\end{table}

\begin{table}[]
	\centering
	\caption{AUC value in CTR prediction.}
	\label{tab:ctr}
	\begin{tabular}{c|ccccc}
		\hline
		Dataset   & CKE   & NFM   & Ripple & KGAT  & KCAN           \\ \hline
		MovieLens & 0.530 & 0.809 & 0.903  & 0.841 & \textbf{0.907} \\ \hline
		Last-FM   & 0.850 & 0.904 & 0.889  & 0.905 & \textbf{0.923} \\ \hline
		Yelp      & 0.915 & 0.908 & 0.917  & 0.922 & \textbf{0.937} \\ \hline
	\end{tabular}
\end{table}

\begin{figure}
	\centering
	\includegraphics[scale=0.63]{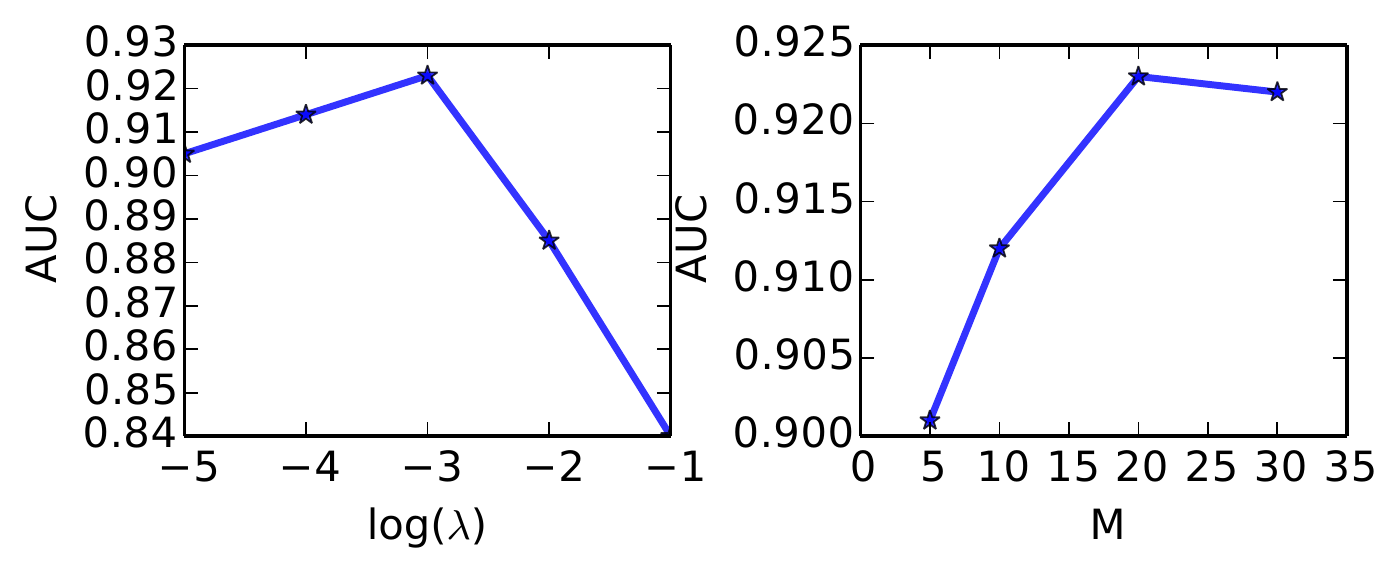}
	\caption{Left: AUC w.r.t. the log value of the weights of $L2$ normalization $\log(\lambda)$. Right: AUC w.r.t. the fixed-size number of neighbors $M$ in LCSAN.}
	\label{fig:PS_l2_S}
\end{figure}

The performance comparison results of the top-K recommendation and CTR prediction are presented in Table \ref{tab:topk} and \ref{tab:ctr}, respectively. The observations are illustrated as follows:
\begin{itemize}
	\item Our proposed KCAN achieves significant improvements over the baselines on all the datasets in both top-K recommendation and CTR prediction. It demonstrates the effectiveness of our proposed method KCAN. Moreover, KCAN outperforms the KGAT in all experiments, and it indicates the effectiveness of conditional attention for refining the knowledge graph to capture the local user preference.
	\item CKE is the worst one in most cases. It demonstrates that directly using the knowledge graph as a regularization can not make full use of the knowledge graph. Besides, the NFM outperforms CKE because it preserves the second-order similarity implicitly by the input of the cross feature.
	\item KGAT achieves better performance than CKE, NFM, Ripple in Last-FM and Yelp datasets. It verifies the effectiveness of propagating information in the knowledge graph to preserve global similarity. 
	\item In MovieLens dataset, CKE, NFM, KGAT have poorer performance than the performance in the other two datasets. The reason is that MovieLens dataset has very few users and items, and a large number of knowledge graphs, it is more difficult to exploit knowledge graph effectively in this dataset. However, Ripple achieves a good performance in this dataset. Ripple uses paths from an item in a user’s history to a candidate item, and the path-based methods can capture the local preference. It indicates the importance of preserving local preference instead of the whole knowledge graph, especially in a massive knowledge graph. Moreover, the improvements of our method KCAN over Ripple may prove that the way we distill the knowledge graph is better than than the way using multiple paths. 
\end{itemize}

\subsection{Study of KCAN (RQ3)}
In this part, we evaluate how different parts of KCAN affect the performance. To study their respective effects, we compare our KCAN with three variants: (1) $\rm{KCAN_{w/o\ LC}}$. It removes the LCSAN layer from KCAN. (2) $\rm{KCAN_{w/o\ GK}}$. It removes the KAGCN layers and uses the knowledge graph embedding as the input of LCSAN. (3) $\rm{KCAN_{w/o\ both}}$. It removes both the LCSAN and KAGCN. The result is shown in Table \ref{tab:study}. The $\rm{KCAN_{w/o\ both}}$ performs the worst among the three variants, it demonstrates the two layers, KAGCN and LCSAN, are beneficial for knowledge-aware recommendations. This conclusion has been double verified since the three variants are worse than the KCAN. Besides, the $\rm{KCAN_{w/o\ GK}}$ outperforms the $\rm{KCAN_{w/o\ LC}}$, which demonstrates the necessity of distilling and refining the knowledge graphs.

\begin{table}[]
	\centering
	\caption{Effects of KAGCN, LCSAN on top-K recommendation.}
	\label{tab:study}
	\resizebox{\columnwidth}{!}{%
	\begin{tabular}{c|c|cccc}
		\hline
		\multicolumn{2}{c|}{Dataset}         & $\rm{KCAN_{w/o\ LC}}$ & $\rm{KCAN_{w/o\ GK}}$ & $\rm{KCAN_{w/o\ both}}$ & KCAN           \\ \hline
		\multirow{2}{*}{MovieLens} & Hit@10  & 0.648            & 0.654            & 0.632              & \textbf{0.668} \\ \cline{2-2}
		& NDCG@10 & 0.355            & 0.364            & 0.348              & \textbf{0.365} \\ \hline
		\multirow{2}{*}{Last-FM}   & Hit@10  & 0.712            & 0.750            & 0.698              & \textbf{0.771} \\ \cline{2-2}
		& NDCG@10 & 0.442            & 0.478            & 0.433              & \textbf{0.506} \\ \hline
		\multirow{2}{*}{Yelp}      & Hit@10  & 0.772            & 0.785            & 0.735              & \textbf{0.810} \\ \cline{2-2}
		& NDCG@10 & 0.497            & 0.503            & 0.469              & \textbf{0.527} \\ \hline
	\end{tabular}}
\end{table}

\subsection{Parameter Sensitivity (RQ4)}

\begin{figure}
	\centering
	\includegraphics[scale=0.65]{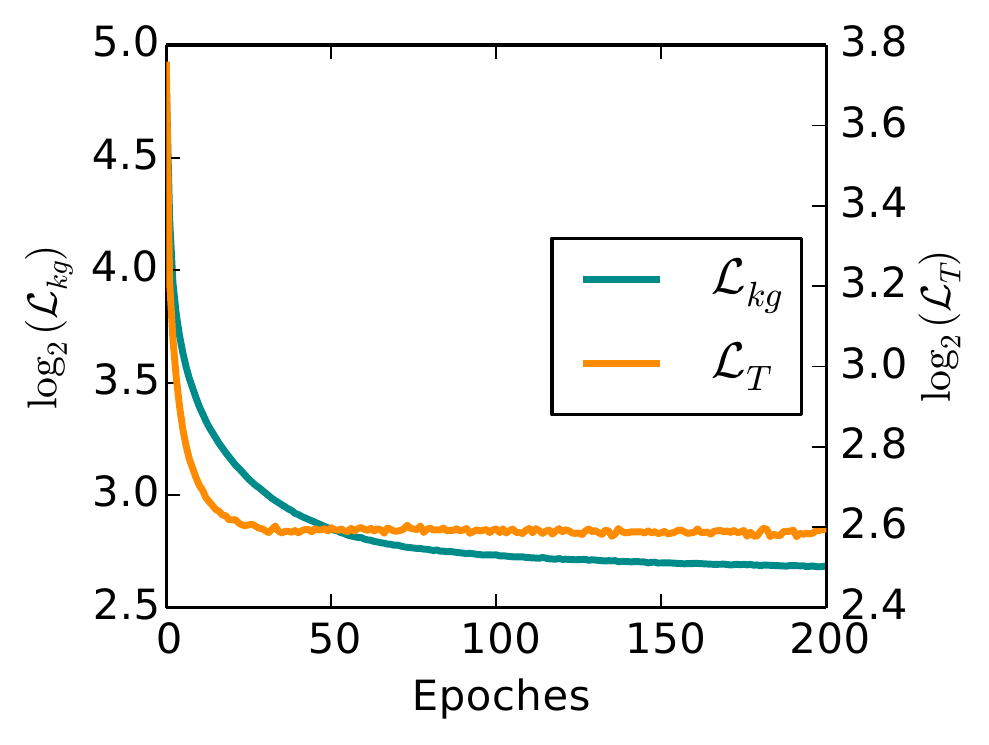}
	\caption{The loss w.r.t the number of epoches.}
	\label{fig:PS_loss}
\end{figure}

In the section, we evaluate the scalability and how different settings of hyper-parameters affect the performance of KCAN. Especially, we evaluate the effect of the weights of $L2$ normalization $\lambda$ and the fixed-size number of neighbors $M$ in LCSAN.
Besides, we also measure the convergence of the alternative optimization in this part.
For brevity, we only report the AUC results with Last-FM datasets, and similar trends can be observed on the other datasets.
\subsubsection{Weight of $L2$ normalization $\lambda$.}
We show how the weights of $L2$ normalization $\lambda$ affect the performance in Figure \ref{fig:PS_l2_S} Left. The $\lambda$ varies from $\{10^{-1}, 10^{-2}, 10^{-3}, 10^{-4}, 10^{-5}\}$. When $\lambda$ increases from $10^{-5}$ to $10^{-3}$, the performance is improved, demonstrating that the $L2$ normalization can avoid overfitting to some extent. When $\lambda$ increases from $10^{-3}$ to $10^{-1}$, the performance becomes poorer. It makes sense since the normalization is much larger than the losses which need to be optimized in this situation.

\subsubsection{The fixed-size number of neighbors $M$.}
In the target-specific sampling step, we sample a fixed-size set of neighbors to distill the knowledge graph and reduce the training time. We vary the fixed-size number of neighbors $M$ from $\{5, 10, 20, 30\}$. The result is shown in Figure \ref{fig:PS_l2_S} Right. We can see that the performance raises firstly when the fixed-size number of neighbors $M$ increases. This is reasonable because a larger $M$ can embody more information in the subgraphs. After $M$ larger than 20, the curve is relatively stable. It demonstrates that our algorithm is not very sensitive to $M$. Besides, we can find that a small $M$, such as 20, can also achieve a relatively good result.

\subsubsection{The convergence of the alternative optimization. }
In the KCAN model, an alternative optimization is used to optimize the loss of knowledge graph embedding $\mathcal{L}_{kg}$ and the loss of target prediction $\mathcal{L}_T$, respectively. In this part, we measures the convergence from the experiment by plotting the loss curve. The number of epoches varies from 1 to 200. The result is shown in Figure \ref{fig:PS_loss}. We can see that the two curves descends very quickly, indicating the efficiency of the framework. About 100 epoches, the two loss are relatively stable, demonstrating our KCAN can converge fast. Besides, we can find that the loss curve of $\mathcal{L}_t$ has very tiny shakes. It is reasonable because the process of sampling target-specific subgraph will introduce sampling bias. 

%% file: Conclusion.tex
\section{Conclusion}
In this paper, we investigate the problem of incorporating the knowledge graph into the recommender system. To address the knowledge graph distillation issue and knowledge graph refinement issue, we propose a novel knowledge-aware graph convolutional network model named \emph{Knowledge-aware Conditional Attention Networks} (KCAN). The framework consists of two main modules, the Knowledge Graph Distillation 
module, and the Knowledge Graph Refinement module. We propagates embedding with knowledge-aware attention in a recursive way to capture the global similarity of entities. After that, a subgraph sampling strategy is designed to distill the knowledge graph based on the attention weight of the KAGCN. Also, the LCSAN propagates personalized information on the sampled subgraph based on local conditional attention for refining the knowledge graph. Benefitting from the two layers, the proposed KCAN can effectively preserve topology similarity and distill and refine the knowledge graph at the same time. Extensive experiments on three real-world scenarios are conducted to demonstrate the effectiveness of our framework over several state-of-the-art methods. In the future, we aim to further reduce the time complexity of KCAN. Another interesting direction is to make the recommendation more explainable. 